\documentclass[letterpaper, 10 pt, conference]{ieeeconf}  % Comment this line out if you need a4paper

\IEEEoverridecommandlockouts                              % This command is only needed if 
\usepackage{todonotes}

\title{\LARGE \bf
When Robots Interact with Groups, Where Does the Trust Reside?
}

\author{Ben Wright$^{1}$ and Emily C. Collins$^{2}$ and David Cameron$^{3}$% <-this % stops a space
\thanks{*This work was not supported by any organization}% <-this % stops a space
\thanks{$^{1}$Ben Wright is a Research Engineer at the Kostas Research Institute for Homeland Security of Northeastern University
        {\tt\small b.wright@northeastern.edu}}%
\thanks{$^{2}$Emily C. Collins is an Associate Research Scientist at the Institute of Experiential Robotics at Northeastern University
        {\tt\small e.collins@northeastern.edu}}%
\thanks{$^{3}$ David Cameron is a Lecturer in the Information School at the University of Sheffield
        {\tt\small d.s.cameron@sheffield.ac.uk}}%
}

\begin{document}

\maketitle
\thispagestyle{empty}
\pagestyle{empty}
\setcounter{footnote}{3}

%%%%%%%%%%%%%%%%%%%%%%%%%%%%%%%%%%%%%%%%%%%%%%%%%%%%%%%%%%%%%%%%%%%%%%%%%%%%%%%%
\begin{abstract}

As robots are introduced to more and more complex scenarios, the issues of trust become more complex as various groups, peoples, and entities begin to interact with a deployed robot. This short paper explores a few scenarios in which the trust of the robot may come into conflict between one (or more) entities or groups that the robot is required to deal with. We also present a scenario concerning the idea of repairing trust through a possible apology.

\end{abstract}
%%%%%%%%%%%%%%%%%%%%%%%%%%%%%%%%%%%%%%%%%%%%%%%%%%%%%%%%%%%%%%%%%%%%%%%%%%%%%%%%

\section{Introduction \& Motivation}

% -A person would be able to address problems in real time and understand consequences
% -The robot doesn’t have rights, and so the issue is who is responsible? The people involved could be culpable, but the situation is not transparently addressing that. 
% So if a robot can make a semi-autonomous decision, who is culpable if that decision goes wrong? Or has a poor outcome? 

% If I had you a cup of hot tea I am able to regulate the hotness of the tea to ensure I don’t put a customer or myself at risk - as soon as that tea gets itself a semi autonomous layer with which it can dispense itself to customers, I need to know what happens to me, my job, my security, if the robot makes a mistake with the tea’s hotness. 

A longstanding and multifaceted core-challenge within the development of social robotics and their use in social settings relates to the robots' trustworthiness \cite{naneva2020systematic}. While there is a burgeoning literature covering many diverse aspects of trust within HRI, such as robots gaining trust \cite{robinette2013building}, losing and restoring of trust \cite{salem2015would,cameron2021effect}, addressing users' over-trusting a robot \cite{ullrich2021development}, much of the attention in research examines trust in terms of simple dyadic interaction \cite{Nam2020-nh}\footnote{This is not peculiar for trust as a topic within HRI, but rather could be seen as illustrative of the overall field's attention}. 

Even within a `simple' dyadic scenario, variations in social contexts may influence people's trust towards a robot and determine what constitutes trustworthy behaviour \cite{holthaus2021does}. The social context may further derive from people's understanding of who `owns responsibility' for the robot and their motivations, thereby bringing additional agents into potentially \textit{any} current social HRI scenario \cite{cameron2021user}. Moreover, emerging work exploring trust in contexts of larger social groups shines a light on these additional social complexities for a robot to navigate; for example, recent empirical research indicates that trust \textit{within} a group can be shaped via a robot mediating human-human interactions \cite{birmingham2020can} or making its own social disclosures \cite{strohkorb2018ripple}. 

For a robot to determine trust from a group, a simple summation of the respective trust of each individual within the group might not be appropriate; where multiple (potentially concurrent) users are interacting with a system, there may be many approaches to determining trust holistically, let alone optimising for this \cite{leister2012ideas}. Similarly, building trust with one group could potentially diminish trust with another. Given this, ought a robot to simply maximise overall trust, or are there alternatives to take and if so, how may a robot determine this calculus? Research considering how a robot could, or should, potentially navigate dynamic, social, group interaction contexts, can illuminate HRI issues which will arise in real-world contexts, where robots have been deployed alongside humans engaging in their daily life, providing future HRI research directions. %I want to say here "and not waste time with low level dyadic research that may prove to be a dead-end when it comes to real social HRI" but can't think of an elegant way to put it. -- I gave it a go! 
In this paper, we present three scenarios in which a hypothetical social robot must address issues relating to trust across multiple agents and identify appropriate actions across competing demands for trust. 

\section{Scenarios}

For the sake of simplicity, we keep the same premise across the scenarios of a hypothetical \emph{barista robot}, (well beyond state-of-the-art, e.g., \cite{sung2020untact}). This type of service-oriented social robot is envisioned to be able to 
\begin{itemize}
    \item [i] interact with customers
    \item [ii] interact with co-workers
    \item [iii] have the capabilities of functioning as a barista.
\end{itemize}

The setting for the scenarios is chosen as a familiar illustration of co-existing and competing interests within the service industry, an area identified as high interest in enrollment of social robotics \cite{blocher2021ai}. Recent research highlights the dynamic social environment and careful navigation through social interactions that may be expected of barista robots in HRI scenarios with multiple people \cite{hedaoo2019robot}. As identified in their work, social violations such as breaching privacy, can have meaningful impact on how people view the robot and could potentially affect human-human interaction.

While such challenges may also be ones that people face in everyday human-human interaction, we consider the specific requirements necessary for a robot to appropriately navigate these, as both meaningful steps in developing models for trust in robotics, and as illustrative of the complex (often unstated) decisions humans make in such circumstances. Challenges particular to a robot may include its inability to make decisions in real time, using the same level, and complexity, of \textit{a priori} information accessible to humans making equivalent decisions. 

Our setting presents the barista robot as being a member of one group (the service staff) in relation to two other groups (customers and managers). Specifically, the scenarios serve to highlight potential issues for a robot in navigating working \textit{with} service staff \textit{for} a manager \textit{towards} customers. The overarching question is not one of \textit{how can the robot build trust}, but \textit{with whom should the robot seek to build trust}.

\subsection{Overheard Gossip}

This first scenario discusses the conflict in trust between teammates and the company/owner that deployed the robot.

\begin{quote}[Scenario 1]
The robot, working on its tasks, overhears co-workers discussing an upcoming opportunity to unionise their workforce.
\end{quote}

%1. And the most extreme then, given we are now thinking about who has placed our social Barista Robot in this coffee shop - let's say the robot overhears unionisation talk - does it tell the boss to whom it is more closely aligned, or stick with its team mates, the human workers? Is the trust of the robot here maintained by a robot that WILL report that to the owner/manager, or is trusting this robot maintained cause the robot is held to confidentiality to its human team mates, and we must consider the issue of which country this team and their robot are in? (and where the data goes - union leader (e.g. in Germany maybe) or straight to the employer - perhaps in USA?), 

There are a few potential actions the robot could take upon hearing these conversations. It could tell the boss/owner about the conversations happening. It could also inform its teammates that it will tell the owner. It could do nothing, reporting to neither group. A number of these actions revolve around the issue of deciding where privacy and first-order trust reside for the robot. Is the robot an extension of company policy verbatim or is the robot's adaptability to increase its own particular team's efficiency the higher priority?

\subsection{An Accident and Mea Culpa}

This second scenario discusses how trust management and repair might work between teammates and customers.

\begin{quote}[Scenario 2]
    After spilling a cup of coffee on a customer, the robot apologises to the customer but is still yelled at for incompetence within earshot of co-workers.
\end{quote}

%2. Apology to the customer after a spill - how does that affect team mates? - so how do the human team mates see the robots? does its mistake bring the humans closer together hating their boss who put it there, or just hate the robot for making them all look incompetent, or are the sympathic of a thing that simply gets in the way but is cute? 

This scenario focuses less on the next action of the robot, but more on the question of changes in trust from this event. How does this scenario impact the human teammates attitudes towards the robot? Would this engender sympathy from them? Would this make them think the robot is incompetent and a nuisance to their team? While trying to repair trust from a mistake with a customer, does the apology build trust with other customers who overhear it?  

\subsection{An (Intentionally) Mislabelled Order}

This third scenario discusses the conflict in trust between teammates, customers, and social norms.

\begin{quote}[Scenario 3]
    The robot is handed a completed order to deliver/announce to the customer base, however the robot notes that the written name on the order can be interpreted as offensive to the customer and is not the name the customer gave when ordering.
\end{quote}

%3. A human teammate writes the name of a customer done that might be construed as controversial, e.g. it says pig/it says something offensive, and the robot has to say the name.. or not, does it back up the team mate or the customer?
%trust code switch - the robot making the proactive decision to not say the name..? 

In this last scenario, the robot once again has a few possible actions it can take. The straight forward thing might be to say the problematic name that its teammate wrote down. Alternatively, another option might be to call out what the order was, instead of the name, reducing the risk of a possible conflict. This all depends on the types of social norms that the robot executes (and its expected to execute). Should the robot help its teammates perform mild forms of protest? Or should the robot's first-order trust reside with the customer's comfort and satisfaction?

\section{Discussion}

These simple scenarios provide some context for the complexities involved in trust management that various deployed robotic systems will encounter. If robots are to become truly interactive in non-trivial environments, then managing various levels of trust will be inevitable. 

Non-dyadic trust, or group trust, may become especially important to navigate. These scenarios ask a number of questions around privacy sharing, group trust building, and issues surrounding group-alignment. Finding situations in which a robot will have to choose one group over another - and how that could affect the underlying trust of the excluded group - is not far fetched. As they are group level decisions, it is also integral to begin looking at trust properties of groups over merely collections of individuals.

These thought experiments are intended to highlight issues with social robots that will be able to make decisions in real-world scenarios. It is not a case of, \emph{should} a robot make decisions, but rather, an acknowledgement of the need to address what needs to be in place \emph{when robots do} make decisions in the real-world. What social and legal infrastructure needs to be in place to ensure everyone in these situations is clear on who is responsible for the actions and outcomes of the advanced technology in the work-place? Considering who trusts whom in these human-robot-manager situations is one practical starting point to answering such a question.

% We want to make clear, why robots are different to people here, and our stance on the future of advanced social robotics - It is not a case of ‘robots should not make decisions’, it is a case of ‘infrastructure should be in place to make sure everyone is clear on who is responsible given the use of advanced technology in the workplace. Who’s job is at risk? What can be done to ensure the workers have the least stressful work situation to work in - which happens when unpredictable events are mitigated for, and everyone clearly knows their rights. CITE ourselves again. 
%We cannot make robots accountable for the decisions they are making, it is the person who is responsible for the robot who will be responsible... 

\bibliographystyle{IEEEtran}
\bibliography{References}

\end{document}